\documentclass[pdflatex,sn-mathphys]{sn-jnl}

\jyear{2021}%

\theoremstyle{thmstyleone}%
\theoremstyle{thmstyletwo}%

\theoremstyle{thmstylethree}%

\begin{document}

\title[Current state of AI]{Current State of Community Driven Radiological AI Deployment in Medical Imaging}

\author*[1]{\fnm{Vikash} \sur{Gupta}}\email{gupta.vikash@mayo.edu}
\author[1]{\fnm{Barbaros Selnur} \sur{Erdal}}
\author[2]{\fnm{Carolina} \sur{Ramirez}}
\author[3]{\fnm{Ralf} \sur{Floca}}
\author[4]{\fnm{Laurence} \sur{Jackson}}
\author[5]{\fnm{Brad} \sur{Genereaux}}
\author[5]{\fnm{Sidney} \sur{Bryson}}
\author[6]{\fnm{Christopher P} \sur{Bridge}}
\author[7]{\fnm{Jens} \sur{Kleesiek}}
\author[7]{\fnm{Felix} \sur{Nensa}}
\author[8]{\fnm{Rickmer} \sur{Braren}}
\author[9]{\fnm{Khaled} \sur{Younis}}
\author[10]{\fnm{Tobias} \sur{Penzkofer}}
\author[11]{\fnm{Andreas Michael} \sur{Bucher}\textsuperscript{11}}
\author[5]{\fnm{Ming Melvin} \sur{Qin}}
\author[5]{\fnm{Gigon} \sur{Bae}}
\author[12]{\fnm{Hyeonhoon} \sur{Lee}}
\author[13]{\fnm{M. Jorge} \sur{Cardoso}}
\author[13]{\fnm{Sebastien} \sur{Ourselin}}
\author[13]{\fnm{Eric} \sur{Kerfoot}}
\author[5]{\fnm{Rahul} \sur{Choudhury}}
\author[1]{\fnm{Richard D} \sur{White}}
\author[14]{\fnm{Tessa} \sur{Cook}}
\author[5]{\fnm{David} \sur{Bericat}}
\author[15]{\fnm{Matthew} \sur{Lungren}}
\author[5]{\fnm{Risto} \sur{Haukioja}}
\author[4]{\fnm{Haris} \sur{Shuaib}}

\affil*[1]{\orgdiv{Center for Augmented Intelligence in Imaging}, \orgname{Mayo Clinic}, \orgaddress{\state{Florida}, \country{USA}}}

\affil[2]{\orgdiv{Center for Intelligent Imaging}, \orgname{University of California}, \orgaddress{\city{San Francisco}, \country{USA}}}

\affil[3]{\orgdiv{Division of Medical Image Computing}, \orgname{German Cancer Research Center (DKFZ)}, \orgaddress{\city{Heidelberg}, \country{Germany}}}

\affil[4]{\orgdiv{Clinical Scientific Computing, Medical Physics}, \orgname{Guy's and St' Thomas' NHS Foundation Trust}, \orgaddress{\city{London}, \country{UK}}}

\affil[5]{\orgdiv{NVIDIA Inc}, \orgaddress{\city{Santa Clara}, \country{USA}}}

\affil[6]{\orgdiv{Department of Radiology}, \orgname{Massachusetts General Hospital}, \orgaddress{\city{Boston}, \country{USA}}}

\affil[7]{\orgdiv{Institute of AI in Medicine}, \orgname{University Medicine Essen}, \orgaddress{\city{Essen}, \country{Germany}}}

\affil[8]{\orgdiv{Institute for Radiology}, \orgname{School of Medicine, Technical University}, \orgaddress{\city{Munich}, \country{Germany}}}

\affil[9]{\orgname{Philips Inc.}}

\affil[10]{\orgdiv{Department of Radiology}, \orgname{Charité - Universitätsmedizin}, \orgaddress{\city{Berlin}, \country{Germany}}}

\affil[11]{\orgname{University Hospital}, \orgaddress{\city{Frankfurt}, \country{Germany}}}

\affil[12]{\orgdiv{Biomedical Research Institute}, \orgname{Seoul National University Hospital}, \orgaddress{\country{Republic of Korea}}}

\affil[13]{\orgdiv{School of Biomedical Engineering \& Imaging Sciences}, \orgname{King's College London}, \orgaddress{\city{London}, \country{UK}}}

\affil[14]{\orgdiv{Perelman School of Medicine}, \orgname{University of Pennsylvania}, \orgaddress{\country{USA}}}

\affil[15]{\orgname{Microsoft Corporation}}



\abstract{
\textbf{Background:}
Artificial Intelligence (AI) has become commonplace to solve routine everyday tasks. Because of the exponential growth  in medical imaging data volume and complexity, the workload on radiologists is steadily increasing. We project that the gap between the number of imaging  exams and the number of expert radiologist readers required to cover this increase will continue to expand, consequently, introducing a demand for tools that improve the efficiency with which radiologists can comfortably interpret these exams. 

\textbf{Methods:} AI has been shown to improve efficiency in medical image generation, processing, and interpretation, and a variety of such AI models have been developed across research labs worldwide. However, very few of these, if any, find their way into routine clinical use, a discrepancy that reflects the divide between AI research and successful AI translation. To address the barrier to clinical deployment, we have formed the Medical Open Network for Artificial Intelligence (MONAI) Consortium, an open source community which is building standards for AI deployment in hospitals, and developing  tools and infrastructure to facilitate their implementation. This paper represents several years of weekly discussions and hands-on problem solving experience by groups of industry experts and clinicians. 

\textbf{Findings:} We identify barriers between AI model development in research labs and subsequent clinical deployment and propose solutions. Our report provides guidance on processes which take an imaging AI model from development to clinical implementation in a healthcare institution. We discuss various AI integration points in a clinical radiology workflow. We also present a taxonomy of radiology AI use cases. 

\textbf{Interpretation:} Through this paper, we intend to educate the stakeholders in healthcare and AI (AI researchers, radiologists, imaging informaticists, and regulators) about cross-disciplinary challenges and possible solutions. 

\textbf{Funding:} MONAI is an open-source project. Funding for authors was made available through the respective affiliated institutional salaries.}




\maketitle
\section*{Abbreviations}
\begin{enumerate}
\item[] AI = Artificial Intelligence
\item[] MONAI = Medical Open Network for Artificial Intelligence
\item[] IT = Information Technology
\item[] DICOM = Digital Imaging and Communications in Medicine 
\item[] PACS = Picture Archiving and Communication Systems
\item[] VNA = Vendor Neutral Archive
\item[] MIMPS = Medical Image Management and Processing Systems
\item[] HIS = Hospital Information System
\item[] RIS = Radiology Information System
\item[] CIS = Clinical Information System
\item[] EMR = Electronic Medical Record 
\item[] HER = Electronic Health Record
\item[] DICOM-SR = DICOM Structured Report 
\item[] DICOM-SEG = DICOM Segmentation
\item[] CDE = Clinical Data Element
\item[] CDSS = Clinical Decision Support System
\item[] ED = Emergency Department
\item[] CT =Computed Tomography
\end{enumerate}

\section{Introduction}
There are multiple well-recognized applications of Artificial Intelligence (AI) in healthcare. Radiology has become the leading focus of healthcare-AI research, and there has been an exponential rise in the number of related publications and AI-enabled devices approved by the United States Food and Drug Administration (FDA)\cite{muehlematter2021approval}. However, the use of AI in a clinical workflow has received little attention from the industry and the research community. Ebrahimian, et al. found that most approved AI/Machine Learning (ML) models are solely image processing focused (the remaining representing a range)\cite{ebrahimian2022fda}. According to an online worldwide survey conducted by the authors and anecdotal personal evidence, one of the key challenges to scaling the transformation of medical imaging operations related to AI is the lowering of technical barriers to deploying, integrating, and orchestrating AI applications relative to clinical workflow patterns. 

The need to develop standard operating procedures for AI and Radiology led to the formation of the Medical Open Network for Artificial Intelligence (MONAI) Consortium, a nonprofit organization led by industry experts and academic physicians and scientists\cite{monai}. The Consortium is divided into various working groups, each concentrating on different aspects of imaging-AI. This report from the Consortium outlines the various considerations viewed as important to an AI deployment within standard Radiology clinical workflows. The goal of MONAI is to present an end-to-end labeling, training and deployment framework for AI applications in the hospitals. There is however a monai model-zoo where people can share their model that can be consumed for either inference or fine-tuning. MONAI provides a framework and guidelines for model sharing, so that the models can be downloaded and deployed with minimum effort. It is also intended to educate stakeholders in the healthcare-AI community about implementation challenges and to discuss interdisciplinary terminology and solutions. Through this paper, we would like to discuss the considerations involved in taking a trained model (in a research lab) through the deployment process and clinical adoption. We will also present some use cases deployed at various contributing site clinical environments.

\subsection{Barriers to Incorporating Imaging-AI in Clinical Practice}
One of the most essential goals in deploying imaging-AI is streamlining the radiologist experience. The AI processing must seamlessly integrate with existing clinical workflows, where imaging impacts downstream decisions like surgery, interventions, and drug prescriptions. These new AI based workflows should not interfere with established routine practices. Radiologists must also be trained in the use of AI-augmented processes without significant interruption of their normal activities. AI tools should have configurable endpoints which can be integrated into diverse medical facility/healthcare system infrastructures across multiple vendors and protocols. An optimally deployed AI-based infrastructure should be indistinguishable from the existing IT environment and require minimal or no training of new radiologist users. Nevertheless, the introduction of such tools requires the fostering of trust among the users (e.g., radiologists and Radiology technical personnel) and benefactors (e.g., patients).

Generic AI applications (e.g., news feeds, shopping recommendations) have become ubiquitous in our lives; employing a centralized infrastructure, the associated user experience adapts to include such AI applications. Related novel support-tools and vocabulary (e.g., tap, drag) have entered our common vernacular only in the last 15 years. This behavior modification was driven largely by smartphone and computer manufacturers.

On the other hand, such centralized operations are not effective in medical facilities and healthcare systems, primarily due to the nonuniform nature of deployment infrastructures. This inhomogeneity is related to differences in acquisition protocols, radiologist workflows, data management procedures, and IT architectures. The evolution of Radiology-AI workflows is complicated, as different components of this system progress at different paces and in different directions, yet are required to be interconnected at all times. A good analogy can be drawn with NASA's space shuttle program. Throughout its lifetime, different technologies had been developed that advanced science. As with the progression of NASA's space shuttle program, all newly introduced technologies (e.g., computer controls) had to be backwards-compatible with previous versions and, at the same time, remain future-proof. Similarly, AI-based clinical workflows need to be retrofitted to existing imaging technology as well as remain flexible enough to work in a rapidly changing IT infrastructure. In addition, novel AI models should be able to consume DICOM images and produce outputs compatible with available Radiology systems (e.g., image viewers). Accordingly, AI workflows should support data standards to promote system interoperability.

\subsection{Inter-System Communication and Interoperability}
Medical facilities and healthcare systems rely on many software applications to meet diverse clinical, research, educational, and business needs. Such applications track patient-care across disciplines (including Radiology, Surgery, Internal Medicine subspecialties, Pathology \& Laboratories, Pharmacy), support inpatient monitoring, and facilitate procedure scheduling, billing, and much more. These applications must be interoperable so that information is not repeatedly entered, managed, or siloed in a single system or department. Recognizing this need more than 30 years ago, the National Electrical Manufacturers Association (NEMA), American College of Radiology (ACR), and Health Level 7 (HL7), collaborated to develop standards, such as Digital Imaging and Communications in Medicine (DICOM) and HL7, now used worldwide. Recently, the Fast Healthcare Interoperability Resources (FHIR)\cite{HL7} standard based on ordinary web technologies (e.g., JSON, XML, REST) has emerged as a valuable data-exchange standard in the healthcare sector. Together with established ontologies, nomenclatures, and classifications (e.g., SNOMED, LOINC, RadLex), FHIR-based profiles enable a semantically interoperable exchange of machine-readable data. 

It is important that an AI deployment system addresses these standards and provide a ubiquitous interface with a wide range of systems. With the rise of AI applications in healthcare, the need for system interoperability has become even more critical as potentially many AI applications will be deployed. Integrating the Healthcare Enterprise (IHE) has defined Profiles that organize and leverage the aforementioned integration capabilities. IHE Profiles contain specific information about diverse clinical needs. While developing AI applications and defining the integration points and workflows, IHE Profiles should be consulted as a  gold standard for deployment and interoperability. While creating the MONAI Deploy tools, IHE Profiles were consulted, and we showed the interoperability capabilities of our tools on two different occasions: IHE Connectathon 2022 (Atlanta) and Imaging AI in Practice (IAIP) organized by RSNA, 2022 (Chicago). During this activity, we deployed end-to-end solutions showing interoperability between different commercial vendors and our open-source solutions.

\section{Current State of AI in the Radiology workflow}
Research bridging AI and Radiology must consider data parameters (e.g., access, quality), patient interests (e.g., privacy, ethics, liability), coding \& billing, and system maintenance; these must be addressed to enable widespread adoption of imaging-AI. While most medical facilities and healthcare systems maintain their data centers and servers due to data-security concerns, AI tools are increasingly implemented as cloud-based inference systems \cite{tragaardh2020recomia, retico2021enhancing, egger2022studierfenster}. As healthcare entities adopt cloud infrastructures, AI deployments need to be increasingly concerned about data security. In addition to traditional security layers, cloud-based systems must also be protected against adversarial attacks that craft errant inputs forcing model errors. Finlayson, et. al have discussed how adding adversarial noise to data may lead to inaccurate diagnoses\cite{finlayson2018adversarial}. Both AI developers and their healthcare customers must be aware of such vulnerabilities.

With an increasing emphasis on federated learning and model sharing as alternatives to data transmission/pooling, the hazards of sharing AI models\cite{finlayson2021clinician} from clinical and dataset shifts\cite{quinonero2022dataset} become relevant. Meaningful changes to the patient population or data-acquisition methods (i.e., a dataset shift) can expose the bias in an AI model\cite{gupta2020democratizing, rockenbach2022automatic}. We believe that building tools that are universal and flexible will allow smaller practices, as well as medical facilities and healthcare systems, to participate in local AI-model refinement rather than solely relying on larger institutions to train and share models. In federated learning, the general idea is that there  are labelled datasets at different hospitals and there is a central server which holds a model which is simultaneously trained on this dataset that is distributed across sites. Thus the infrastructure is built around simultaneous training. However, if we want to implement a framework which supports continuous Federated Learning and train a federated model on a relatively “large” dataset, we need to implement a deployment and labeling framework that is compatible with the routine clinical workflow. We would like to re-iterate the fact that it is not always feasible to ask a radiologist to perform annotation task outside their clinical duties because of time constraints. 

Our claims to further develop the tools for clinical AI deployment are corroborated by the report titled "Brilliant AI Doctor"\cite{wang2021brilliant} and the need to improve rural patient care. While AI is often touted as a means of providing affordable healthcare to the masses, we believe the reality is very different, as is outlined in a recent WIRED article titled "AI can help diagnose some illness, if your country is rich". The WIRED authors argue that although AI development might be supportable by larger and prosperous institutions with enough radiologists and technical personnel for training and deployment of AI models, such infrastructure is neither present nor affordable in many rural healthcare settings. However, suppose we deploy the AI model on the routine clinical environment. In that case, smaller hospitals' radiologists can seamlessly adjudicate the AI results and fine-tune or develop a new model without large overhead costs.

According to Wang, et. al, even tertiary hospitals in China rely on very basic printable electronic health record (EHR) systems\cite{wang2021brilliant}. In addition, during the COVID-19 pandemic, many hospitals in India and China resorted to personal communication tools like WhatsApp and WeChat to help connect patients with doctors; critical medical information discussed on these platforms is not recorded in the EHR, and this valuable patient data is consequently lost. Therefore, for many practice settings, effective models cannot be developed in the absence of appropriate data-capture tools. Because of the potential for dataset shift, models developed at other healthcare institutions may not perform well locally. Hence, the dilemma where models cannot be trained due to insufficient data, and data are not collected because there are no models being trained. Thus, we need to consider novel AI solutions for regions with limited resources to meaningfully deliver the AI promise of quality, affordable, state-of-the-art healthcare. 

\section{AI Integration Points in Healthcare Infrastructure}
Interoperability in healthcare is a necessity as there are many applications from different vendors required in any given workflow. These applications use standards like HL7\cite{HL7} and DICOM\cite{DICOM} alongside IHE\cite{ihe} profiles that describe how these standards can be used to connect applications together. Understanding the interoperability in a given workflow is the first step when considering AI integration touchpoints.
Figure 1 shows a basic radiology workflow, modeled after IHE Scheduled Workflow\cite{Schwork} that shows an order being generated, images being acquired from the patient, these images being reviewed, and a report being generated and sent back to the clinician for review.
\begin{figure}
    \centering
    \includegraphics[scale=0.5]{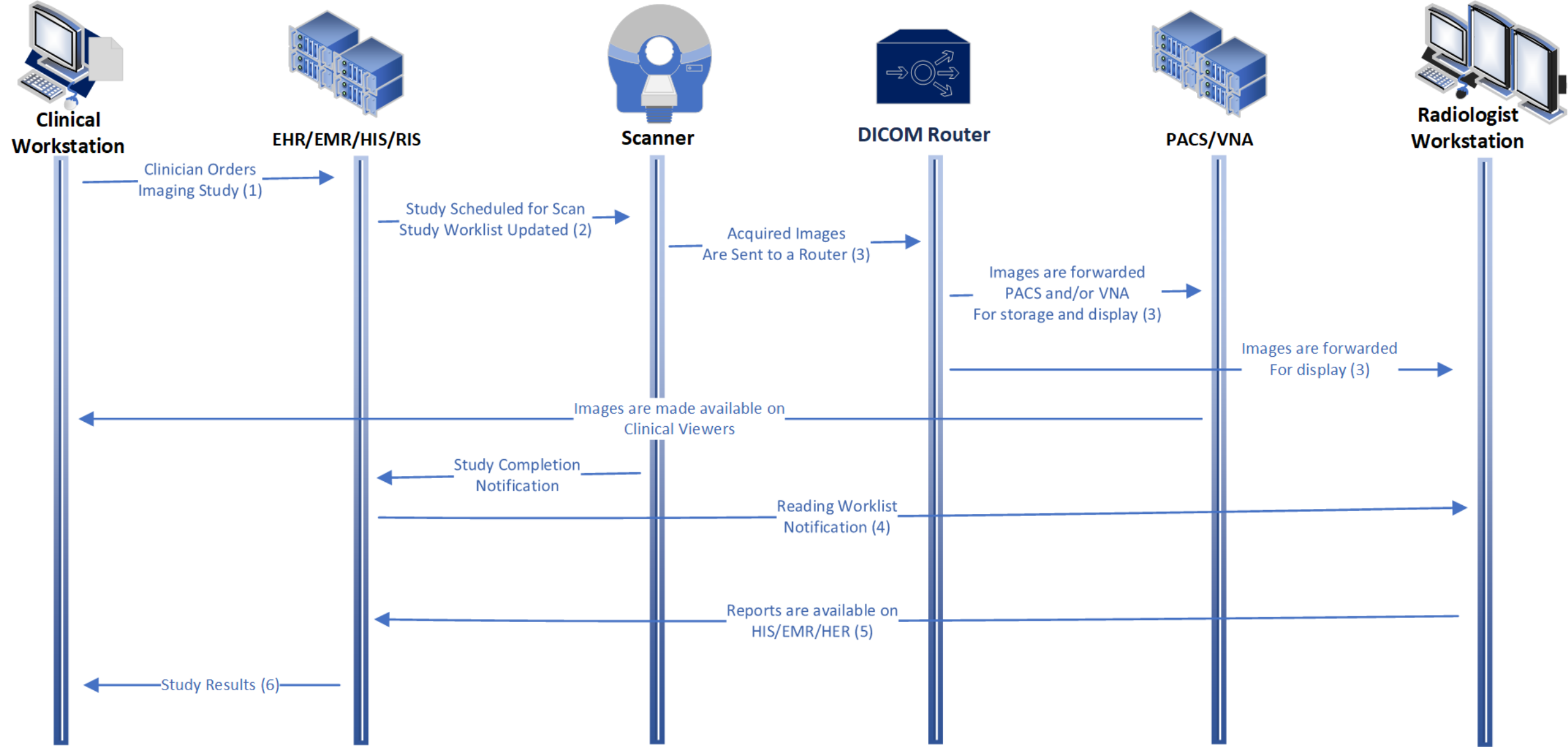}
    \caption{A basic Radiology workflow with core components and messaging interfaces}
    \label{fig:my_label1}
\end{figure}
This is a common radiology workflow, with the following notes:
\begin{enumerate}
    \item When a clinician orders an imaging examination in the HIS/RIS, they may be guided by a CDSS to ensure its appropriateness. Depending on the clinical setting, the order may contain a clinical-status priority code (i.e., 'stat', 'urgent', etc.)
    \item Once the patient examination is scheduled for a date and location, an entry is created on the "study worklist" of the scanner (or another imaging device). In some instances, an entry is also created on a "protocoling worklist", where a radiologist determines the imaging techniques to be used (e.g., scanning details, contrast-agent type/amount/administration route) during the diagnostic imaging study or image-directed procedure.
    \item Once the examination is completed, images are reconstructed into a human-interpretable format and sent to a DICOM-router to be forwarded to the appropriate destinations, including a PACS and/or VNA for management or storage. Once the organized images (original and/or post-processed) are ready to be viewed by the radiologist, the examination description appears on the radiologist's "reading worklist". 
    \item Radiologists review the examination images on their diagnostic viewer and dictate their interpretation (typically into a voice recognition system). 
    \item The dictated report is sent to the HIS/RIS. If actionable critical and/or non-critical findings are identified, radiologists may invoke additional workflows to alert the ordering clinician, along with issuing the final examination report.
    \item Final examination reports become available in the HIS/EHR, along with the images in the PACS or clinical viewers.
\end{enumerate}

\begin{figure}
    \centering
    \includegraphics[scale=0.5]{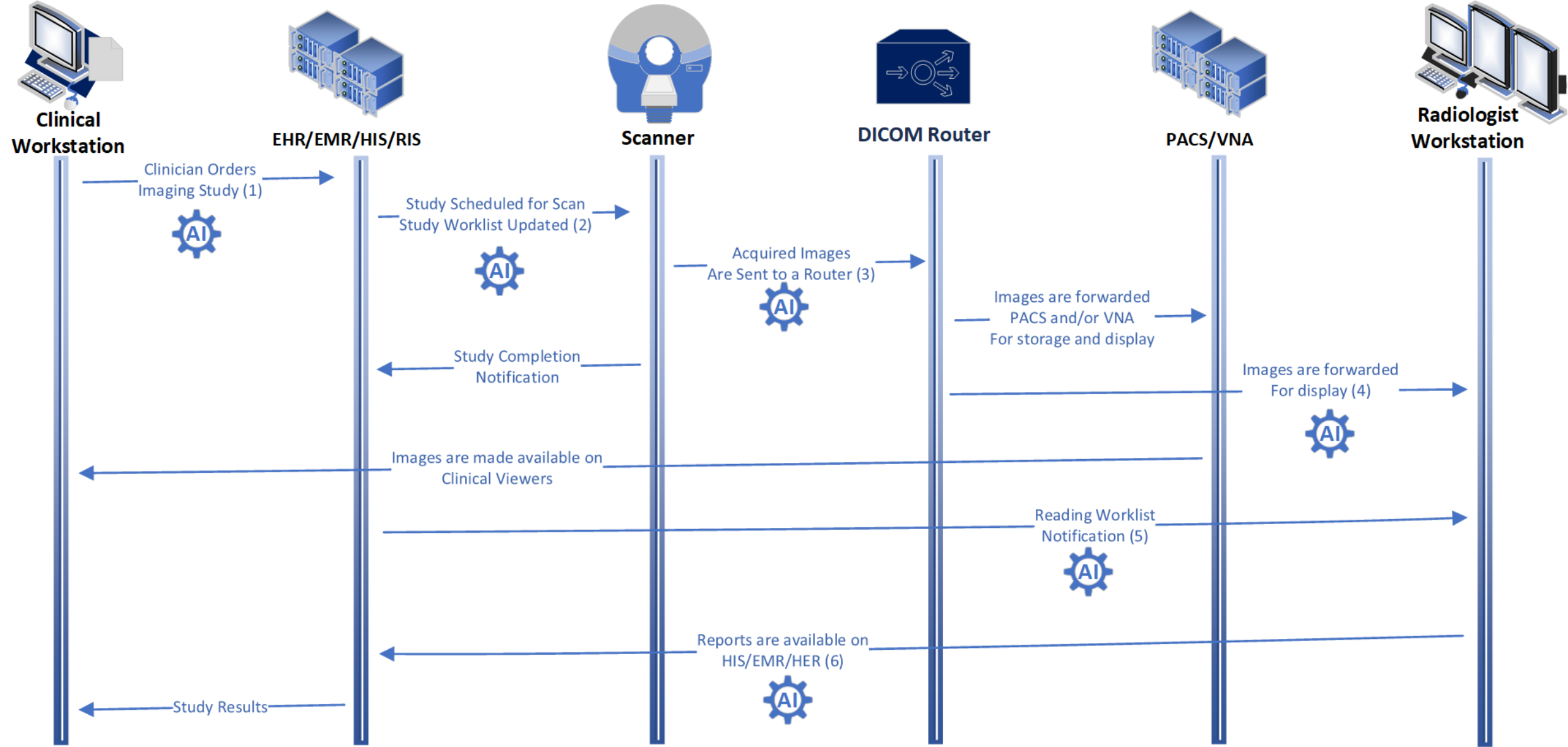}
    \caption{A basic Radiology workflow with core components and messaging interfaces, and possible imaging AI integration points.}
    \label{fig:my_label2}
\end{figure}

The opportunities for using AI across the workflow leverages interoperability to make it feasible. The types of workflows are described in the IHE AI in Imaging White Paper \cite{IAIP}, and profiles AI Workflow for Imaging (AIW-I) \cite{AIWI}  and AI Results (AIR) \cite{AIR} describe the interoperable boundaries. Some examples of using AI in Figure 2 include:
\begin{enumerate}
    \item Using AI to make recommendations as to the types of procedures that should be ordered, based on the patient's condition and record.
    \item Using AI to make recommendations for the type of protocol to be used on the scanner.
    \item Using AI to post-process the image, identify QA issues prior to the patient leaving the department, and prepare classifications and segmentations in advance of the radiologist reader.
    \item Using AI to include insights alongside the images in the radiologist display.
    \item Using AI to include emergent insights for consideration of the ordering physician.
    \item Using AI to pre-populate the radiologist report with draft insights to be considered by the radiologist.
\end{enumerate}

For all of these workflows to interoperate at scale, DICOM is used for the worklist, used in the transmission of the images, and capture and present the insights to the destination systems. HL7 is used for transmitting orders and reports between systems. 

While there are early stages of cloud-based PACS solutions, for most healthcare organiztions today, the norm is still on-prem deployments.  On the AI space,  there are many vendor examples which utilize cloud based implementations, however since most PACS systems are on-prem, this brings hybrid (on-prem + cloud based) implementations into picture.  In these scenarios vendors deliver solutions through either client systems (which connect to their cloud servers), or by delivering their results to destination workstations through proxy servers operating across firewalls. MONAI Deploy \cite{monaisdk} enables producing containerized (Docker containers) which can be executed on-prem or on-cloud which can be decided based on given site’s architectural needs. While PyTorch is the preferred deep learning framework,  TensorFlow is supported as well.  For efficient task and resource management MONAI Deploy offers a Task Management systems which can we employed (if needed) for managing containers. For facilitating standards based (DICOM, HL7) communication, MONAI Deploy offers an Informatics Gateway solution. All aforomention components can be deployed and utilized using tools such as Docker Compose or through Kubenetes based implementations.

\section{A Taxonomy of Imaging-AI Use-Cases}
In this section, we present categories and scenarios where imaging-AI could be helpful.

\subsection{Emergency Diagnostics}
Time is of the essence in some emergency situations (e.g., stroke, cardiac arrest, trauma), when the Emergency Department (ED) must await imaging results to determine the most suitable next steps in patient care. For example, when a patient has been in a motorcycle accident, the ED often orders emergency CT imaging to detect possible skull fractures and brain hemorrhage \cite{prevedello2017automated}; this could be expedited with imaging-AI. Alternative would be waiting for radiologists to go through a reading worklist in sequential order, before arriving most emergent cases.  

\subsection{Integrated Diagnostic Radiology Examination Planning}
Currently, the scheduling of a new imaging examination and the review of prior imaging data for protocoling of that examination is typically performed separately and asynchronously by different participants (schedulers/technologists vs physicians, respectively). With the help of imaging-AI, these two key components of diagnostic Radiology-examination planning could be made to be both concurrent and complementary, leading to optimized prospective examination planning, including enhanced scanner and protocol selection.

\subsection{Opportunistic Screening}
The term "opportunistic screening" refers to the application of imaging-AI technology to pixel data to improve wellness, prevent disease, assess risk, or detect asymptomatic disease. The opportunistic concept focuses on enhanced screening for silent conditions or risk factors that are incidental to the primary indication for the imaging examination. Among the most promising opportunistic-screening use-cases is the training of a model to detect and quantify Coronary Artery Calcium (CAC) for cardiovascular risk stratification on routine chest CTs performed with high frequency for non-cardiovascular reasons (e.g., interstitial lung disease, trauma, low-dose lung nodule screening) \cite{van2020deep}. Quantification of CAC on non-cardiac-gated chest CTs is complicated by motion-related artifacts and protocol heterogeneity.

Another relevant use-case for opportunistic screening is CT-based body composition analysis, which quantifies liver fat, organ volumes, and muscle loss, among other characteristics \cite{kroll2021assessing, koitka2021fully, kroll2022ct}.

\subsection{Interactive Diagnostics during Image Interpretation}
Radiologists often consult their Radiology colleagues for diagnostic mediation when interpreting challenging or ambiguous cases (e.g., possible cancers, subtle fractures, coronary stenosis in motion-degraded vessels, cancer detection in dense breasts). In the process, the work of both radiologists is interrupted, and department workflow is slowed. The alternative of not seeking such peer-consultation, but rather ordering additional diagnostic evaluations can prolong patient assessment, delay treatment initiation, and/or lead to additional risks and expenses. Hence, an AI-based "second-opinion" support system could be highly beneficial to both radiologists and patients.

\subsection{Large-Scale AI Model Validation}
Radiologists increasingly collaborate with AI researchers to identify new areas of research and model development. Such collaborations yield large-scale statistical analyses to identify trends and gather insights about disease progression and risk factors; this can stimulate novel radiomic/radiogenomic-based diagnostics. For example, there are ongoing efforts in the neuroimaging community to identify imaging biomarkers for Alzheimer's \cite{frisoni2010clinical} and Parkinson's disease \cite{weingarten2015neuroimaging}. Consortiums such as the Alzheimer's Disease Neuroimaging Initiative (ADNI), Parkinson's Progression Markers Initiative (PPMI), UK Biobank, and The Cancer Imaging Archive (TCIA) are facilitating the development of AI for imaging-driven precision medicine. 

While most current AI applications focus on a direct radiological application, the use of AI on medical imaging data is much broader in scope. AI can be used to predict disease progression (e.g. predicting cognitive decline and atrophy in patients suffering from Alzheimer's disease \cite{grueso2021machine}, therapeutics inferencing (e.g. finding disease subtypes and optimal drugs in patients with multiple sclerosis \cite{eshaghi2021identifying}), and several downstream interventions, such as surgical guidance \cite{vercauteren2019cai4cai}, electrode placement \cite{neumann2019toward}, among others. Such systems need to be integrated with the clinical workflow beyond standard radiological endpoints, often requiring bespoke user interfaces and integration with the EHR. These use cases, while important, are outside of the scope of this article, as it is primarily focusing on radiological applications. 

Large-scale AI model needs to be validated by rigorous evaluation and standardized reporting on AI applications in healthcare. As stated by Panch et al. there has long been a gap between the expectation created from impressive small-scale research evaluations on medical AI applications and a relative sparsity of distribution of similar applications in the real-world clinical pathways \cite{panch2019inconvenient}. There are numerous reasons to explain this gap, but a very important point is the lack of well-established, common infrastructures to assist the training and evaluation, and distribution of AI models between healthcare institutions. These kinds of infrastructures can only be derived from large-scale collaborations, as described in section 5.4. Thorough evaluation on large-scale representative collaborative networks, are irreplaceable in order to test AI for its validity, as has been illustrated in detail by Ghassemi et al. in their analysis of explainable AI methods and the fallacy of false hope that can be brought about by the current state of available explainability methods \cite{ghassemi2021false}. In conclusion, reliability should rather be derived from evaluations on large scale, representative datasets.

\section{Real World Use Cases}
Several institutions represented in the MONAI Consortium have begun the implementation of AI applications within their medical-imaging workflows. These institutions include Mayo Clinic, University of California San Francisco (UCSF), University of Pennsylvania (UPenn), Massachusetts General Brigham (MGB), Guy's and St Thomas' Hospital (GSTH), National Health System (NHS), and German Cancer Research Center (DKFZ). The MONAI consortium has various working groups, such as MONAI Core, Deploy, and Label. The MONAI consortium and group of developers collaborate to develop tools, which are used to build the solutions for the use cases presented in this section. We present representative real-world use cases that can be built, reproduced, and/or deployed using current tools available in MONAI.

\subsection{Clinical Decision Support Systems}
Many of the aforementioned institutions have CDSS in place to aid clinicians in selecting appropriate imaging examinations for their patients. The majority of CDSS depend on the American College of Radiology's Appropriateness Criteria \cite{acr}. In addition, most of these institutions also employ dictation systems with AI capabilities.

However, in the following sections, the use-cases discussed will focus on imaging-AI operating between the placement of an examination order and the production of a final radiological report.

\subsection{Pediatric Bone-Age Determination}
Bone age is a valuable metric of skeletal maturity in pediatric and adolescent patients. It is normally performed by an experienced radiologist who manually compares the bones on a frontal X-ray of the hand and wrist against a decades-old atlas. This is a time-consuming process that must be performed by a radiologist with related expertise and yet remains prone to interpreter variability. 

To address this variability, GSTH investigators integrated a commercial AI bone-age application into its clinical workflow [Figure 2]. The bone-age application \cite{bone} was deployed according to the vendor specification with dedicated hardware running the service, and users were required to manually send imaging studies to the DICOM node when calculating bone age. 

\begin{figure}[!t]
    \centering
    \includegraphics[scale=0.9]{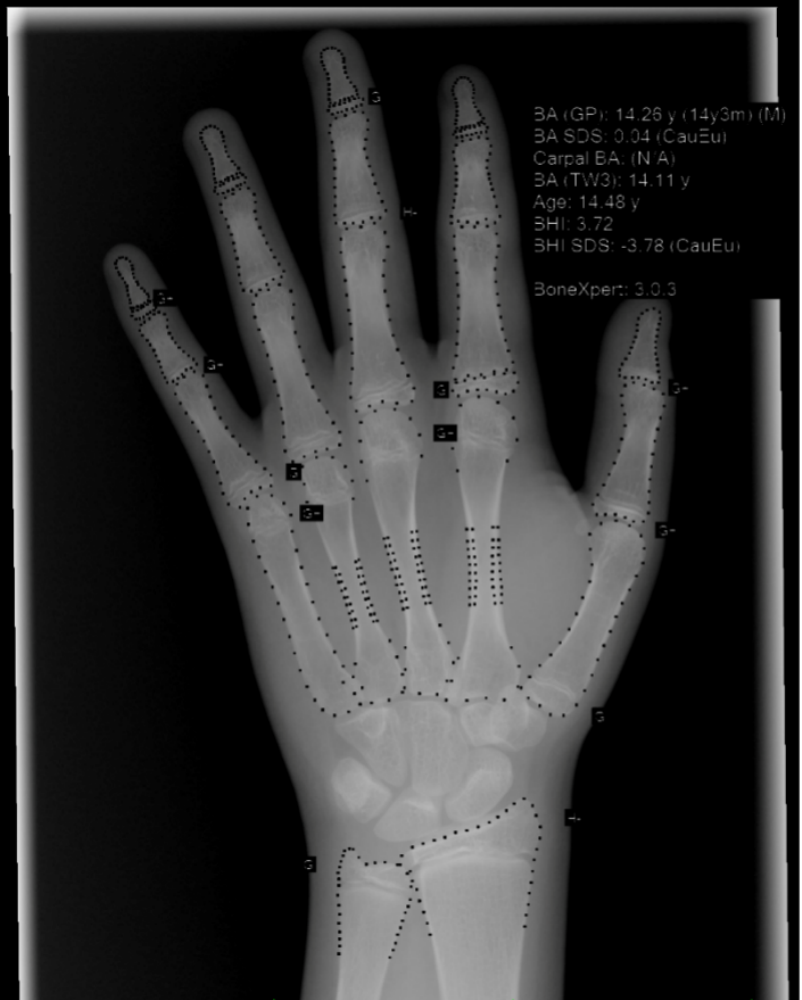}
    \caption{Output from the AI bone-age application showing identified features and quantitative results}
    \label{fig:my_label3}
\end{figure}
Approximately 3 months after deployment, a study was performed to quantify radiologists' time saved by the application. In fact, no significant time savings was found when using the AI bone-age application versus manual measurement by radiologists. The most likely explanation was that the radiologists had to forward the examination data to the AI system when reporting, thereby nullifying any potential efficiency that the AI system itself introduced. This experience illustrates that the management of clinical workflows and the enablement of imaging-AI applications are closely coupled and cannot be considered in isolation. 

\subsection{Brain Tumor Segmentation and Progression Detection from Longitudinal MRI data}
The current standard for assessing treatment response for gliomas is based on changes in 2D cross-sectional tumor measurements of anatomical brain MRI \cite{van2011response, villanueva2017current, upadhyay2011conventional}. However, studies show that 3D volumetric assessments outperform 2D measurements for reliable tumor progression detection \cite{damasceno2022clinical}.

UCSF investigators deployed a ML-based Human-in-the-Loop (HITL) workflow to automatically segment longitudinal low-grade glioma tumors within brain MRI data, compute volumetrics, and generate reports showing tumor-volume changes along several timepoints. This clinical workflow was intended to assist physicians during tumor board meetings in assessing treatment options for low-grade glioma patients. In its design, a patient undergoes a brain MRI and the images are sent to the PACS. If a physician then requests tumor volumetrics for the patient prior to an upcoming tumor board, an analyst selects FLAIR acquisitions from the current examination and five prior studies. These sequences are sent for automatic AI-based segmentation, which outputs a DICOM-SEG object of the tumor ROI and the estimated volume (in ml). These outputs are sent to XNAT, where the analyst can review and correct the ROI as needed, which will trigger automatic re-calculation of tumor volume. A radiologist performs a final review, and upon approval, a report is automatically generated and pushed to the PACS as a DICOM Secondary Capture. 

These interactive and interpretable progression assessments can be presented to radiologists within 10 minutes of DICOM transmission, as shown in Figure 4. Automated segmentation has lower variability in volumetric measurement compared to manual segmentation, and performance matches the gold standard under the same acquisition sequence (GE 3D CUBE) \cite{huang2020volumetric}.

\subsection{RACOON Network}
RACOON unites all university hospitals in Germany in a nationwide infrastructure (https://racoon.network) that enables federated training and application of AI for medical imaging. As part of the Network of University Medicine \cite{netzwerk} the project represents a consortium consisting of DKFZ, all 38 German University Radiology departments, public research institutes, Fraunhofer Mevis, Technical University Darmstadt, and commercial partners (Mint Medical and ImFusion). RACOON's hybrid network architecture enables federated analysis through common nodes at all partner sites (RACOON-NODE) complemented by a secure cloud environment that allows the pooling of datasets and hosts central services (RACOON-CENTRAL). Each node provides a toolset for structured reporting \cite{salg2021reporting}, image annotation, and segmentation, as well as an open-source platform for AI training and automated image analysis (Joint Imaging Platform) \cite{scherer2020joint}.

Two priorities of the project are (1) enabling generation and access to well-structured, quality-controlled, and representative datasets for AI-model development and evaluation, and (2) distribution of AI methods and their federated training. Therefore, the first set of specialized solutions cover the large scale model validation category (section 4.5). However, due to RACOON's versatile architecture and the homogenous hardware and software setup, which is integrated into the local hospital infrastructure of all partners, RACOON will also help to realize other previously mentioned general use-cases (e.g., "scheduled Radiology diagnostics" or "opportunistic screening") 

In an initial specific clinical use-case, a large, structured dataset of COVID and comparable pulmonary diseases was collected across all RACOON sites, along with method developments for quality assurance, image segmentation, outcome prediction, and evaluation of reporting scores. 

\subsection{Detecting MRI-Hazardous Implanted Leadless Electronic Devices on Radiographs}
Particular implanted leadless electronic devices are considered either stringently MRI-conditional or MRI-unsafe, necessitating restriction to basic MRI scanning or patient exclusion from an MRI examination, respectively. Ideally, the implanted device type is documented in a patient's EHR and available to MRI -examination supervisors. Otherwise, radiographs are often used to pre-screen patients for such devices and, accordingly, for their eligibility to undergo standard MRI. Unfortunately, such devices can be easily overlooked or misrecognized on x-rays due to their small sizes, similar appearances, or suboptimal image acquisition (e.g., insufficient contrast between the device and surrounding tissues). This can lead to serious consequences or injury to the patient during MRI, especially at newer high-field strengths (e.g., 7 Tesla).

\begin{figure}
    \centering
    \includegraphics[scale=0.09]{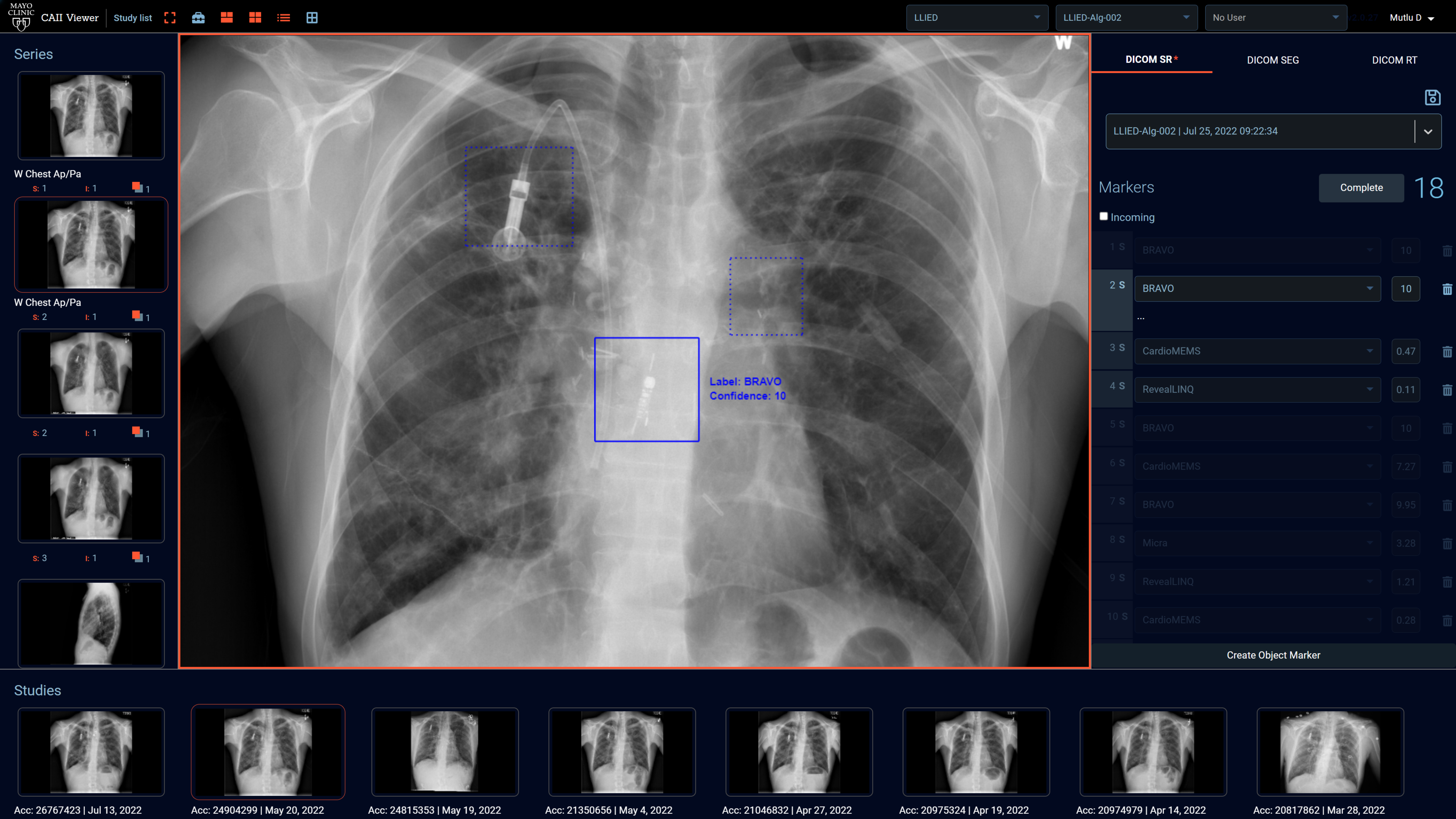}
    \caption{MRI-unsafe device (Bravo esophageal reflux pH capsule) correctly detected and identified (with 10/10 certainty) on chest x-ray by model inference (shown as a solid bounding box) displayed and adjudicated by radiologist on viewer}
    \label{fig:my_label5}
\end{figure}

To address these issues, Mayo Clinic investigators developed, upgraded by continuous learning, and deployed a frontal chest x-ray-based device detection algorithm \cite{white2022pre}. To that end, two cascading AI models were utilized, with the results of the first model (for detection based on a Faster R-CNN) fed into the second model (for identification based on a multi-class CNN). To make this tool operational and usable in the clinical setting, MONAI-based frameworks were employed and a robust "viewer" with back-end data recording was developed, as shown in Figure \ref{fig:my_label5}.

\section{Conclusions}
While AI is often touted as a "sentient" entity that will soon replace physicians (including radiologists), this misconception is far from reality. From the standpoints of implementation and adoption, AI in healthcare and medical imaging is still in its infancy, despite extensive publishing on state-of-the-art AI models. In addition, to our knowledge, there are few AI applications that can yet prove time savings to radiologist workflows. Measuring efficiency would only be possible by implementing standards, such as IHE Standard Log of Events (IHE SOLE) \cite{sole}, in addition to careful user-interface designs to enable users to provide feedback and utilize the results with minimal workflow impact. Hence, the system and the algorithm performances can be captured, logged, and measured objectively. Such efforts by MONAI Deploy are underway, and initial demonstrations of how to provide such measurement mechanisms were demonstrated at IHE Connectathon, 2022 (Atlanta) and RSNA Imaging AI in Practice, 2022 (Chicago) in coordination with the vendor community.

The goal of the MONAI platform is to present an end-to-end labeling, training, and deployment platform for medical imaging. In this report, initial AI-model deployments of clinical use cases at multiple institutions represented in the MONAI Consortium \cite{monai, monaisdk} were highlighted; as a guide for navigating the complex terrain of clinical deployment, a taxonomy of different general use cases was also presented. Most of the perspectives and recommendations in this report emanated from a worldwide survey (\ref{appendix}) by the MONAI Consortium and weekly intensive Consortium-member discussions over the past three years. 

This report represents an effort specifically by the MONAI-Deploy Group of the MONAI Consortium to elucidate the challenges of imaging-AI clinical deployment and to guide deployment architectures and solutions to common deployment issues.  To that end, we delineated the current state of imaging-AI implementation within clinical workflows in Radiology, discussing the potential barriers and suggesting responses. We believe it is reasonable to expect imaging-AI with adequately trained models to aid radiologists in their routine clinical practices; AI is likely to reduce radiologist workloads in the near future. However, to incorporate imaging-AI into Radiology workflows, an infrastructure that efficiently and effectively interfaces with existing operations and infrastructures is essential. If the promise of AI is in making healthcare, including Radiology, more affordable and impactful, such seamless integration is required to deliver on this potential.

As an association, the MONAI Consortium aims to reduce the obstacles in AI-model development and deployment, thereby improving rates of model translation into clinical practice. We believe that by providing tools and APIs that are open-source, community-driven, and adhere to common clinical deployment standards, we can further the adoption of AI in clinical environments. We can reach underserved communities by democratizing the tools needed for seamless AI implementation in a range of healthcare institutions. Our goal is to build an open-source AI workflow where novel models can be trained and deployed with minimum overhead costs to healthcare organizations, thus helping to democratize healthcare AI and reach wide adoption in smaller hospitals in underserved locations.

\section{Acknowledgements}
This is an open-source project. There are no grants or funding specifically associated with this work. All authors for this paper are salaried employees of their respective institutions. Additional funding sources are as follows. Vikash Gupta, Barbaros Selnur Erdal and Richard D. White are funded through the Mayo Clinic Foundation. Ralf Floca is funded from the German Research Society and German Federal Ministry of Education and Research. Brad Genereaux, Sidney Bryson, Ming Melvin Qin, Gigon Bae, Rahul Choudhury, David Bericat, Risto Haukioja are paid employees of NVIDIA Inc.  Jorge Cardoso, Eric Kerfoot and Sebastien Ourselin acknowledge funding from the Wellcome/EPSRC Centre for Medical Engineering (WT203148/Z/16/Z) and Wellcome Flagship Programme (WT213038/Z/18/Z). Jorge Cardoso, Sebastien Ourselin and Haris Shuaib also acknowledge funding from the London Medical Imaging and AI Centre for Value-based Healthcare. Matthew Lungren is a paid employee of Microsoft Nuance. Khaled Younis is an employee of Philips Inc. Tessa Cook wishes to acknowledge grants from Independence Blue Cross, National Institute of Health, Radiological Society of North America, and American College of Radiology. Tobias Penzkofer wishes to acknowledge the German Research Foundation (DFG, SFB 1340/2), German Ministry of Education and Research (01KX2021, 68GX21001A), and European Union Horizon 2020 (Grant No. 952172). We would like to acknowledge contributions from Dr. Marc Modat at the School of Biomedical Engineering \& Imaging Sciences, King's College London to the MONAI community. 

\bibliography{ref}

\pagebreak
\section*{Appendix 1}\label{appendix}

The MONAI-Deploy Group designed a questionnaire and conducted a self-selection survey, initially limited to the institutional partners in this project, and shared via email and social media (twitter, Linkedin and email chains) at a later stage. Details about the questionnaire and survey can be found on the survey website: \url{https://docs.google.com/forms/d/e/1FAIpQLSeuiNLtd-iC-8ZQ_3uK6pyrs3aUVkkLIgrkvv-SlUDcAT6R0w/viewform}

The self-selection nature and limited participation scope does not allow one to make generalizable conclusions. However, the survey design process itself brought up a larger discussion among all the parties involved (e.g., physicians, imaging scientists, informaticists, administrators), regarding the current practical issues in clinical imaging-AI deployments. 

\begin{figure}[!b]
    \centering
    \includegraphics[scale=0.6]{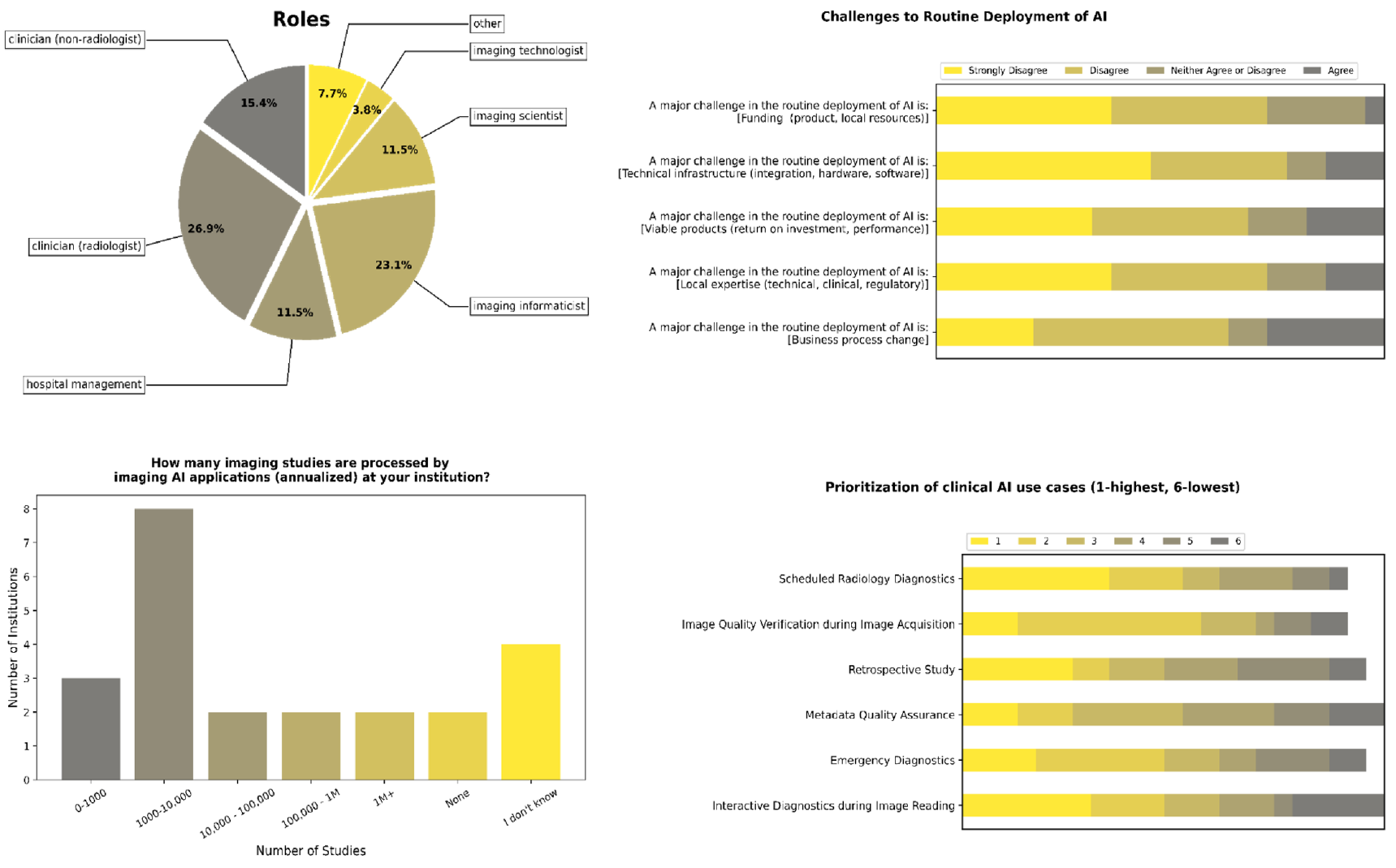}
    \caption{Results of the MONAI-Deploy Group survey showing the: distribution of participants [top-left], volume of workload [bottom-left], barriers to routine deployment [top-right] and prioritization of AI use-cases [bottom-right]}
    \label{fig:summary}
\end{figure}

As shown \ref{fig:summary}, there were 23 respondents to the survey, including 21 from healthcare institutions which were eligible for inclusion; the remaining 2 were ineligible. Several respondents self-identified as multi-professional. Identified professions included radiologists (7), non-radiology physicians (4), imaging informaticists (5), radiographer (1), medical physicist (1), postdoctoral researcher (1), and director of AI \& Data Infrastructure (1).

\end{document}